\documentclass[10pt,letterpaper]{article}

\usepackage{cogsci}
\cogscifinalcopy 
\usepackage{multirow}
\usepackage{graphicx}
\usepackage{pslatex}
\usepackage{apacite}
\usepackage{float} 

\usepackage{dcolumn}
\newcolumntype{d}{D{.}{.}{3}}

\usepackage{booktabs}
\usepackage[colorinlistoftodos,prependcaption,textsize=scriptsize]{todonotes}

\newcommand{\todofk}[1]{\todo[linecolor=red,backgroundcolor=red!25,bordercolor=red]{FK: #1}}

\newcommand{\mc}{\multicolumn}
\newcommand{\word}[1]{\textit{#1}}

\title{Character-based Surprisal as a Model of \\Reading Difficulty in the
  Presence of Errors}
 
\author{\normalfont
  \\
  \begin{tabular}{cc}
  \bf Michael Hahn (mhahn2@stanford.edu) & \bf Frank Keller (keller@inf.ed.ac.uk) \\
  Department of Linguistics, Stanford University & School of Informatics, University of Edinburgh \\
  Margaret Jacks Hall, Stanford, CA 94305, USA & 10 Crichton Street, Edinburgh EH8 9AB, UK \\[1em]
  \bf Yonatan Bisk (ybisk@cs.washington.edu) & \bf Yonatan Belinkov (belinkov@seas.harvard.edu)\\
   Paul G.\ Allen School of Computer Science \& Eng.  &   John A.\ Paulson School of Eng. \& Applied Sciences, \\ University of Washington  &  Harvard University, and Computer Science and Artificial  \\
  185 E Stevens Way NE, Seattle, WA 98195, USA &  Intelligence Laboratory, MIT, Cambridge, MA, USA\\  
  \end{tabular}}
\begin{document}

\maketitle

\begin{abstract}
  Intuitively, human readers cope easily with errors in text; typos,
  misspelling, word substitutions, etc. do not unduly disrupt natural
  reading. Previous work indicates that letter transpositions result
  in increased reading times, but it is unclear if this effect
  generalizes to more natural errors.  In this paper, we report an
  eye-tracking study that compares two error types (letter
  transpositions and naturally occurring misspelling) and two error
  rates (10\% or 50\% of all words contain errors). We find that human
  readers show unimpaired comprehension in spite of these errors, but
  error words cause more reading difficulty than correct words.  Also,
  transpositions are more difficult than misspellings, and a high
  error rate increases difficulty for all words, including correct
  ones.  We then present a computational model that uses
  character-based (rather than traditional word-based) surprisal to
  account for these results. The model explains that transpositions
  are harder than misspellings because they contain unexpected letter
  combinations. It also explains the error rate effect: expectations
  about upcoming words are harder to compute when the context is degraded,
  leading to increased surprisal.

\end{abstract}

\section{Introduction}

Human reading is both effortless and fast, with typical studies
reporting reading rates around 250 words per minute
\cite{Rayner:ea:06}. Human reading is also adaptive: readers vary
their strategy depending on the task they want to achieve, with
experiments showing clear differences between reading for
comprehension, proofreading, or skimming
\cite{Kaakinen:Hyona:10,Schotter:ea:14,Hahn:Keller:18}.

Another remarkable aspect of human reading is its robustness. A lot of
the texts we read are carefully edited and contain few errors, e.g.,
articles in newspapers and magazines, or books. However, readers also
frequently encounter texts that contain errors, e.g., in hand-written
notes, emails, text messages, and social media posts. Intuitively,
such errors are easy to cope with and impede understanding only in a
minor way. In fact, errors often go unnoticed during normal reading,
which is presumably why proofreading is difficult.

The aim of this paper is to experimentally investigate reading in the
face of errors, and to propose a simple model that can account for our
experimental results. Specifically, we focus on errors that change the
form of a word, i.e., that alter a word's character sequence. This
includes letter transposition (e.g.,~\word{innocetn} instead of
\word{innocent}) and misspellings (e.g.,~\word{inocent}).
Importantly, we will not consider whole-word substitutions, nor will
we deal with morphological, syntactic, or semantic errors.

We know from the experimental literature that letter transpositions
cause difficulty in reading
\cite{Rayner:ea:06,Johnson:ea:07,White:ea:08}. However, transpositions
are artificial errors (basically they are an artifact of typing), and
are comparatively rare.\footnote{For example, in the error corpus we
  use \cite{geertzen2014automatic} only 11\% of the errors are letter
  swaps or repetitions, see Table~\ref{tab:error_types}.}  It is not
surprising that such errors slow down reading. This contrasts with
misspellings, i.e., errors that writers make because they are unsure
about the orthography of a word. These are natural errors that should
be easier to read, because they occur more frequently and are
linguistically similar to real words (\word{inocent} conforms to the
phonotactics of English, while \word{innocetn} does not). This is our
first prediction, which we will test in an eye-tracking experiment
that compares the reading of texts with transpositions and
misspellings.

Readers' prior exposure to misspellings might explain why reading is
mostly effortless, even in the presence of errors. The fact remains,
however, that all types of errors are relatively rare in everyday
texts. All previous research has studied isolated sentences that
contain a single erroneous word. This is a situation with which the
human language processor can presumably cope easily. However, what
happens when humans read a whole text which contains a large
proportion of errors? It could be that normal reading becomes very
difficult if, say, half of all words are erroneous. In fact, this is
what we would expect in expectation-based theories of language
processing, such as surprisal \cite{levy_expectation-based_2008}: the
processor constantly uses the current context to compute expectations
for the next word, and difficulty ensues if these expectations turn
out to be incorrect. However, if the context is degraded by a large
number of errors, then it is harder to compute expectations (and they
become less reliable), and reading should slow down. Crucially, we
expect to see this effect on all words, not just on those words that
contain errors. This is the second prediction that we will test in our
eye-tracking experiment by comparing texts with high and low error
rates.

In the second part of this paper, we present a surprisal model that
can account for the patterns of difficulty observed in our experiment
on reading texts with errors. We start by showing that standard
word-based surprisal does not make the right predictions, as it
essentially treats words with errors as out of vocabulary items. We
therefore propose to estimate surprisal with a character-based
language model. We show that this model successfully predicts human
reading times for texts with errors and accounts for both the effect
of error type and the effect of error rate that we observed in our
reading experiment.

\section{Eye-tracking Experiment}

The aim of this experiment was to determine how human reading is
affected by errors in the input. As explained in the introduction, we
expected different error types to affect reading differentially, as
error types can differ in familiarity. In addition, we predicted the
overall number of errors in a text to have an effect on reading
behavior, because a high error rate degrades word context, which is
crucial for computing expectations about upcoming material.

The experiment used a two-by-two factorial design, crossing error type
(transpositions vs. misspellings) with error rate (10\% of all words
contain errors vs. 50\%). Both of these variables were administered as
between-text factors, i.e., we created four versions for each text,
one with 10\% transpositions, one with 10\% misspellings, one with
50\% transpositions, and one with 50\% misspellings.

The two experimental factors were administered within participants,
i.e., all participants read all our texts, each of them presented in
one of the four versions. Versions were distributed across
participants using a Latin square design, so as to ensure that every
version was seen by the same number of participants.

\subsection{Methods}

\subsubsection{Participants}

Sixteen participants took part in the experiment after giving informed
consent. They were paid \pounds{}10 for their participation, had
normal or corrected-to-normal vision, and were self-reported native
speakers of English.

\paragraph{Materials}

We used the materials of \citeA{Hahn:Keller:18}, but introduced errors
into the texts. These materials contain twenty newspaper texts from
the DeepMind question answering corpus~\cite{hermann_teaching_2015}.
Ten texts were taken from the CNN section of the corpus and the other
ten texts from the Daily Mail section.  Texts were comparable in
length (between 149 and 805 words, mean 323) and represent a balanced
selection of topics.  Two additional texts were used as practice
items.

Each text comes with a question and a correct answer. The questions
are formulated as sentences with a blank to be completed with a named
entity so that a statement implied by the text is obtained.  Three
incorrect answers (distractors) are included for each question; these
are also named entities, chosen so that they closely match the correct
answer (e.g.,~if the correct answer is \textit{Minnesota}, then the
distractors are also US states).\footnote{\label{fn:preview}We used
  the no questions preview condition of \citeA{Hahn:Keller:18}, i.e.,
  the questions were shown only after participants had read the whole
  text. The original paper also had a question preview condition, in
  which participants were shown the questions before they read the
  text.}

We introduced errors into the materials of \citeA{Hahn:Keller:18}
following the method suggested by \citeA{Belinkov:18}. These errors
are automatically generated and are either transpositions (i.e.,~two
adjacent letters are swapped) or natural errors that replicate actual
misspellings. For the latter, we used a corpus of human edits
\cite{geertzen2014automatic}, and introduced errors in our
experimental materials by replacing correct words with known
misspellings from our edit corpus. The percentages of different types
of misspellings are listed in Table~\ref{tab:error_types}.
By generating texts with errors automatically we were able to ensure
that both error conditions (transpositions or misspellings) contain
the same percentage of erroneous words for the two error rates (10\%
or 50\% erroneous words).

\begin{table}[tb]
\centering
\begin{small}
\begin{tabular}{@{~}c@{\hspace{10pt}}c@{\hspace{10pt}}c@{\hspace{10pt}}c@{\hspace{10pt}}c@{\hspace{10pt}}c@{~}}
\toprule
phonetics & deletion & swap/repeat & keyboard & insertion & other \\
\midrule
36.2 & 16.7 & 11.0 & 10.5 & 8.3 & 17.3\\
\bottomrule
\end{tabular}
\caption{Percentages of different types of misspellings in the natural
error condition.}
\label{tab:error_types}
\end{small}
\end{table}

\subsubsection{Procedure}

Participants received written instructions, which mentioned that they
would be reading texts with errors. They first went through two
practice trials whose data was discarded. Then, each participant read
and responded to all 20 items (texts with questions and answer
choices); the items were presented in a new random order for each
participant. The order of the answer choices was also
randomized.


In each trial, the text was displayed over one or more pages (max~5,
mean 2.1 pages), where each page contained up to eleven lines with
about 80 characters per line. To get to the next page, and at the end
of the text, participants again had to press a button. After the last
page, the question was displayed, together with the four answer
choices, on a separate page. Participants had to press one of four
buttons to select an answer.

Eye-movements were recorded using an Eyelink 2000 tracker
(SR~Research, Ottawa). The tracker recorded the dominant eye of the
participant (as established by an eye-dominance test) with a sampling
rate of 2000~Hz.
Before the experiment started, the tracker was calibrated using a
nine-point calibration procedure; at the start of each trial, a
central fixation point was presented. Throughout the experiment, the
experimenter monitored the accuracy of the recording and carried out
additional calibrations as necessary.

\subsubsection{Data Analysis}

For data analysis, each word in the text was defined as a region of
interest. Punctuation was included in the region of the word it
followed or preceded without intervening whitespace.  If a word was
preceded by a whitespace, then that space was included in the region
for that word.  We report data for the following eye-movement measures
in the critical regions: \emph{First fixation duration} is the
duration of the first fixation in a region, provided that there was no
earlier fixation on material beyond the region. \emph{First pass time}
(often called gaze duration for single-word regions) consists of the
sum of fixation durations beginning with this first fixation in the
region until the first saccade out of the region, either to the left
or to the right.  \emph{Total time} consists of the sum of the
durations of all fixation in the region, regardless of when these
fixations occur. \emph{Fixation rate} measures the proportion of
trials in which the region was fixated (rather than skipped) on
first-pass reading.  For first fixation duration and first pass time,
no trials in which the region is skipped on first-pass reading
(i.e.,~when first fixation duration is zero) were included in the
analysis. For total time, only trials with a non-zero total time were
included in the analysis.

Due to space limitations, we will only present analyses of the first
pass time and fixation rate data in the remainder of this paper.

\subsection{Results}

\begin{table}[tb]
\centering
\begin{tabular}{@{~}lddd@{~}}
\toprule
& \mc{1}{c}{\multirow{2}{*}{Hahn \& Keller}} & \mc{2}{c}{This experiment}\\
& & \mc{1}{c}{No error}& \mc{1}{c}{Error}\\
\midrule
First fixation & 221.3 &  211.8 & 225.1 \\
First pass     & 260.7 &  242.5 & 265.2 \\
Total time     & 338.0 &  306.9 & 342.1 \\
Fixation rate  &  0.50 &  0.45  & 0.48  \\
\midrule
Accuracy       & \mc{1}{c}{70\%} & \mc{2}{c}{72\%} \\
\bottomrule
\end{tabular}
\caption{Left: per-word reading times, fixation rates, and question accuracies
  in the experiment of \citeA{Hahn:Keller:18}, right: same measures
  for our experiments (same texts, but some of the words contain errors).}
\label{tab:desc_stats}
\end{table}

In Table~\ref{tab:desc_stats}, we present some basic reading measures
for our experiments, and compare these to the reading experiments of
\citeA{Hahn:Keller:18}, which used the same texts, but did not include
any errors (the data is taken from their no question preview
condition, which corresponds to our experimental setup, see
Footnote~\ref{fn:preview}). Even for words with errors, the reading
measures in our experiments are similar to the ones reported by
\citeA{Hahn:Keller:18}. For words without errors, we find slightly
faster reading times and lower fixation rates than
\citeA{Hahn:Keller:18}. Also the accuracy (which can only be measured
on the text level, hence we do not distinguish words with and without
errors) is essentially unchanged. This provides good evidence for the
claim that human readers cope well with errors in text: they take
longer to read words with errors and fixate them more compared to
words without errors, but this this is a comparatively small
effect. Overall, reading times, fixation rates, and question accuracy
are very similar to those found in texts without any errors (such as
the ones used by \citeNP{Hahn:Keller:18}).\footnote{Note that
  participants are not performing at ceiling in question answering;
  our pattern of results therefore cannot be explained by asserting
  that the questions were too easy.}

In the following, we analyze two reading measures in more detail:
first pass time and fixation rate. 
We analyzed per-word reading measures
using mixed-effects models, considering the following predictors:
\begin{enumerate}
\item \textsc{ErrorType}: Does the text contain mispellings ($-0.5$)
  or transpositions ($+0.5$)?
\item \textsc{ErrorRate}: Does the text contain 10\% ($-0.5$) or 50\%
  ($+0.5$) erroneous words overall?
\item \textsc{Error}: Is the word correct ($-0.5$) or erroneous ($+0.5$)?
\item \textsc{WordLength}: Length of the word in characters.
\item \textsc{LastFix}: Was the preceding word fixated ($+0.5$) or not
  ($-0.5$)?
\end{enumerate}
All predictors were centered.  Word length was scaled to unit
variance.  We selected binary interactions using forward model
selection with a $\chi^2$ test, running the R package \texttt{lme4}
\cite{bates-fitting-2015-1} with a maximally convergent random effects
structure. We then re-fitted the best model with a full random effects
structure as a Bayesian generalized multivariate multilevel model
%
%
using the R package \texttt{brms}; this method is slower but allows
fitting large random effects structures even when traditional methods
do not converge. Resulting Bayesian models are shown in
Table~\ref{tab:mixed-models}.  We used the \texttt{brms} default
priors \cite{burkner2017brms}, with four chains with 1000 samples each
(and 1000 warmup iterations).  The $\hat{R}$ values ($\leq 1.01$)
indicated that the models had converged.\footnote{An analogous
  analysis for log-transformed first-pass times led to the same
  pattern of significant effects and their directions.}

\begin{table}[tb]
\begin{tabular}{@{~}l@{}d@{}d@{}l@{~~}d@{}d@{}l@{~}}
\toprule
& \mc{3}{c}{First Pass}& \mc{3}{c}{Fixation Rate} \\
\midrule
(Intercept)                & 248.41 & (6.34) & $^{***}$& -0.16   & (0.12) & $^{}$  \\
\textsc{ErrType}                 & 1.41 & (1.32) & $^{}$    &  0.08   & (0.02) & $^{***}$ \\
\textsc{ErrRate}                 &  7.20 & (1.60) & $^{***}$   & 0.16  & (0.02) & $^{***}$\\
\textsc{Error}                      & 23.77 & (4.12) & $^{***}$ & 0.21  & (0.07) & $^{***}$ \\
\textsc{WLength}                &  22.18 & (2.02) & $^{***}$& 0.83  & (0.04)  & $^{***}$  \\
\textsc{LastFix}                    & 3.10   & (4.18) &         & 0.22 & (0.18) & \\ 
\midrule
\textsc{ErrRate} $\times$ \textsc{LastFix} & 6.71 & (2.77) & $^{*}$    & 0.16  & (0.04) & $^{***}$ \\
\textsc{Error} $\times$ \textsc{LastFix}     &\mc{3}{c}{---}            & 0.26  & (0.10) & $^{**}$  \\
\textsc{WLength} $\times$ \textsc{LastFix} & \mc{3}{c}{---}        & 0.74  & (0.10) & $^{***}$ \\
\bottomrule
\mc{7}{c}{$Pr(\beta < 0)$:  $^{***}<0.001$, $^{**}<0.01$, $^*<0.05$}
\end{tabular}
\vspace{-2ex}
\caption{Bayesian generalized multivariate multilevel models
  for reading measures with maximal random-effects
  structure. Each cell gives the coefficient, its standard deviation,
  and the estimated posterior probability that the coefficient has the
  opposite sign.}\label{tab:mixed-models}
\end{table}

The main effects of \textsc{WordLength} replicate the well-known
positive correlation between word length and reading time (see
\citeNP{demberg_data_2008}, and many others).  We also find main
effects of \textsc{Error}, indicating that erroneous words are read
more slowly and are more likely to be fixated.  The main effects of
\textsc{ErrorRate} show that higher text error rates lead to longer
reading times and higher fixation rates for all words (whether they
are correct or erroneous).  Additionally, we find a main effect of
\textsc{ErrType} in fixation rate, showing that transposition errors
lead to higher fixation rates. This is consistent with our hypothesis
that misspellings are easier to process than transpositions, as they
are real errors that participants have been exposed in their reading
experience.

\begin{figure}[tb]
    \centering
    \includegraphics[width=\columnwidth]{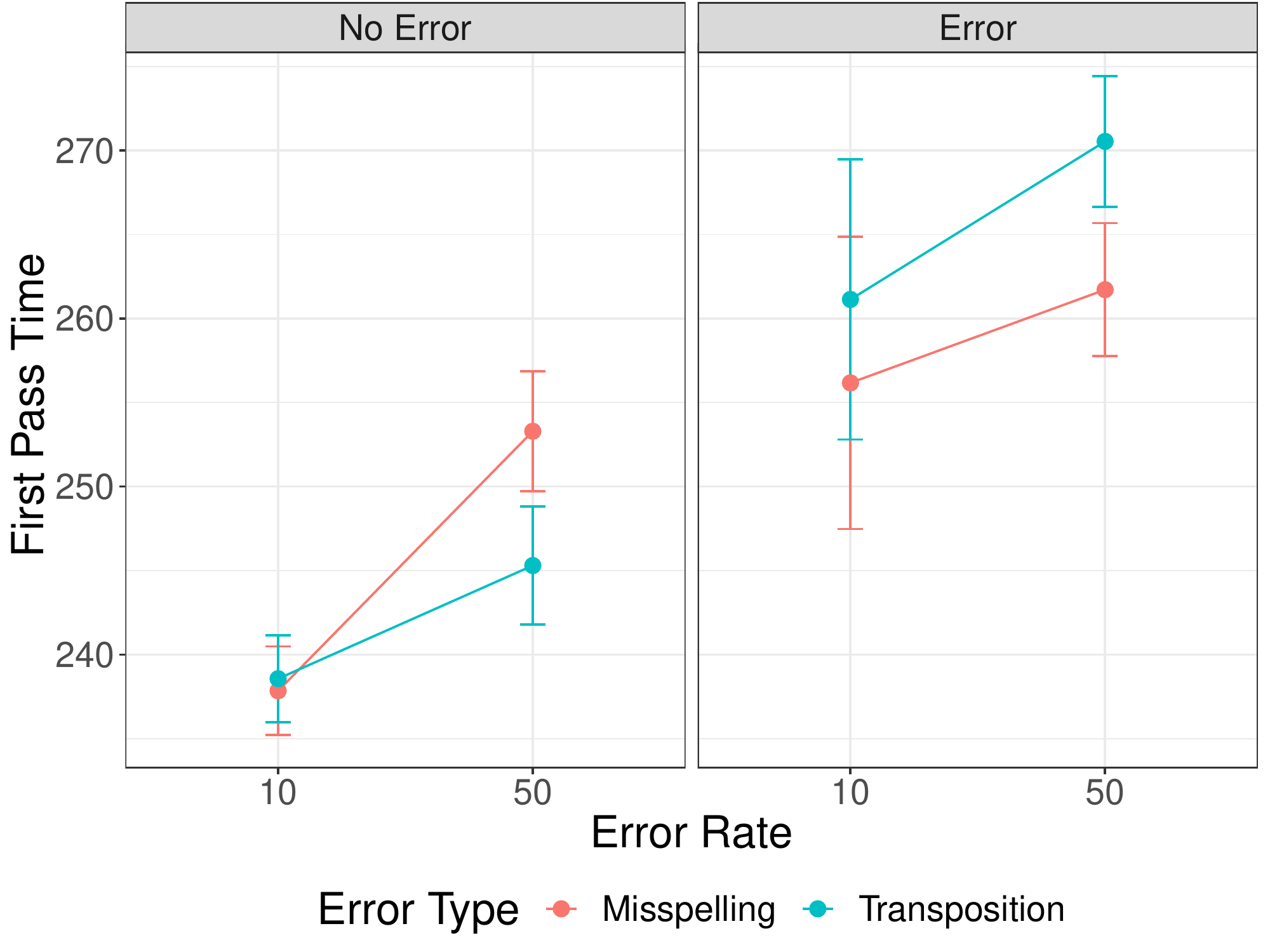}
    \includegraphics[width=\columnwidth]{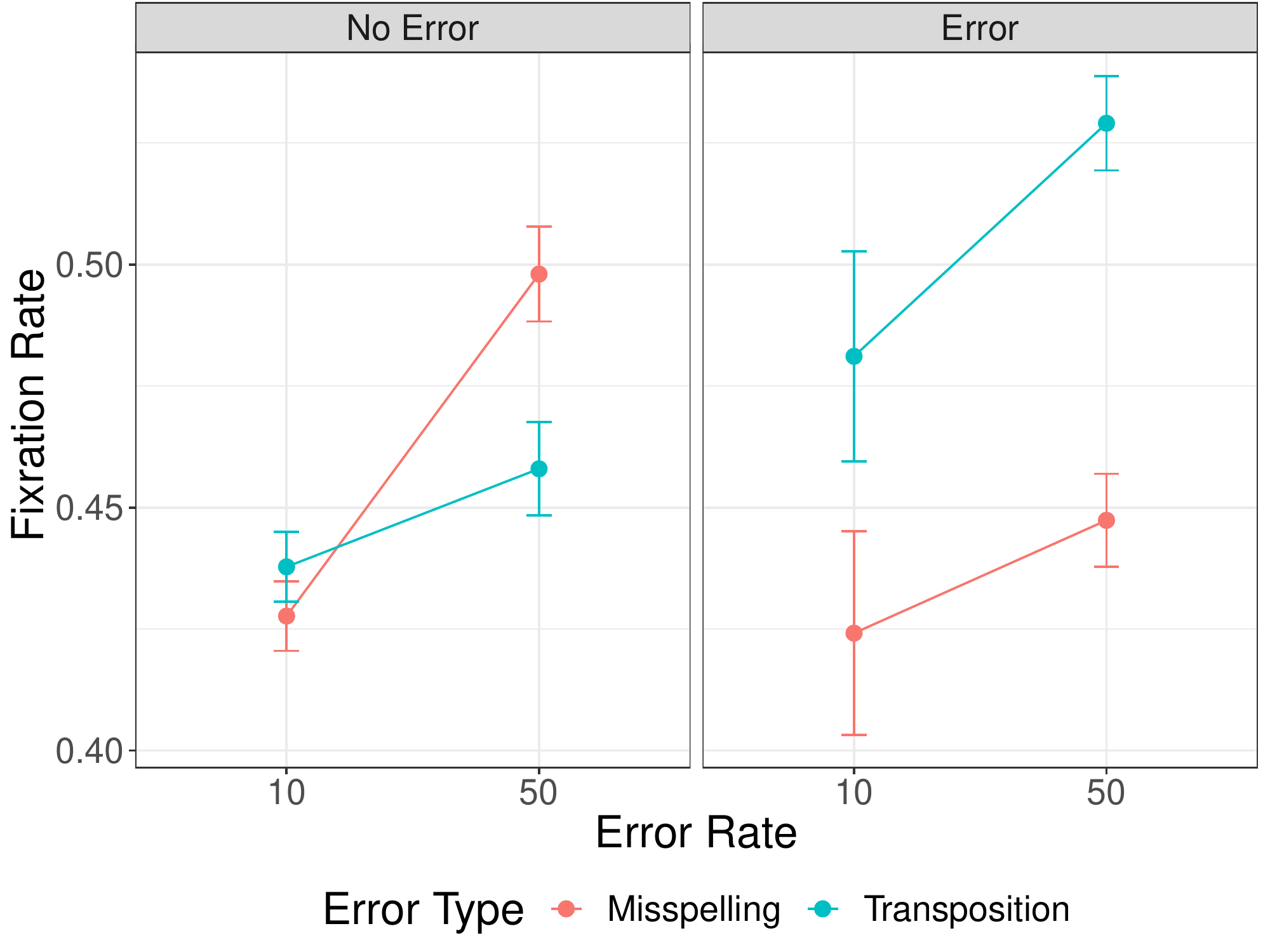}
	\caption{First pass time (top) and fixation rate (bottom)
          when reading texts with transposition errors or misspelling.}
    \label{fig:fp}
\end{figure}

Figure~\ref{fig:fp} graphs mean first pass times and fixation rates by
error type and error rate. The most important effect is that error
words take longer to read and are fixated more than non-error
words. The effect of error rate is also clearly visible: the 50\%
error condition causes longer reading times and more fixations than
the 10\% one, even for non-error words. We also observe a small effect
of error type.


Turning now to the interactions, we found that \textsc{ErrorRate} and
\textsc{LastFix} interact in both reading measures, which indicates
that reading times and fixation rates increase in the high-error
condition if the previous word has been fixated.

Only in fixation rate, there was also an interaction of \textsc{Error}
and \textsc{LastFix}, indicating that fixation rate goes up for error
words if the preceding word was fixated, presumably because of
preview of the erroneous words, which is then more likely to be
fixated in order to identify the error.

For fixation rate, \textsc{WordLength} interacts with
\textsc{LastFix}: longer words are more likely to be fixated if the
preceding word was fixated; again, this is likely an effect of
preview.  While Figure~\ref{fig:fp} seems to suggest an interaction of
\textsc{Error} and \textsc{Error Type}, this was not significant in
the mixed model.

\subsection{Discussion}

We have found four main results: (1)~Erroneous words show longer
reading times and are more likely to be fixated. (2)~Higher error
rates lead to increased reading times and more fixations, even on
words that are correct. (3)~Transpositions lead to an increased
fixation rate compared to misspellings. (4)~Whether the previous word
is fixated or not modulates the effect of error and error rate.

However, it is conceivable that the effects of error and error rate are
actually artifacts of word length. All else being equal, longer
words take longer to read and are more likely to be fixated. So if
error words and non-error words in our texts differ in mean length,
then that would be an alternative explanation for the effects that we
found.  


For transposition errors, error words by definition have the same
length as their non-error versions.  For misspellings, a mixed-effects
analysis with word forms as random effects showed no significant
difference in the lengths of error words and their correct versions
(mean difference $-0.011$, SE $0.029$, $t = -0.393$). Comparing the
erroneous words of the two error types, we found that they differ in
mean length (misspellings 5.44, transpositions 6.06 characters);
however this difference was not significant in a mixed-effects
analysis predicting word length of erroneous words from error types,
with items as a random effect (mean difference $0.015$, SE $0.010$, $t
= 1.449$).  

%
%

\section{Surprisal Model}

Most models of human reading do not explicitly deal with reading in
the face of errors. In fact, reading models that use a lexicon to look
up word forms (e.g.,~to retrieve word frequencies) cannot deal with
erroneous words without further assumptions. We can use the surprisal
model of processing difficulty \cite{levy_expectation-based_2008} to
illustrate this: in its original, word-based formulation, surprisal is
forced to treat all error words as out of vocabulary items; it
therefore cannot distinguish between different types of errors or
between different error rates.

Intuitively, a more fine-grained version of surprisal is required that
computes expectations in terms of characters, not words. In such a
setting, the word \word{inocent} would be more surprising than
\word{innocent} in the same context, but not as surprising as a
completely unfamiliar letter string. In other words, the surprisal of
the same word with and without misspellings or letter transpositions
would be similar but not the same. To achieve this, we can use
character-based language models, which are standard tools in natural
language processing for dealing with errors in the input (e.g.,~the
work by \citeNP{Belinkov:18}, on errors in machine translation).

Crucially, once we have a character-based surprisal model, we can
derive predictions regarding how errors should affect reading. We
predict that transpositions should be more surprising than
misspellings, as they involve character sequences that are unfamiliar
to the model (e.g.,~\word{innocetn} contains the rare character
sequence \word{tn}). Also, we predict that words that occur in texts
with a high error rate are more difficult to read than words in texts
with a low error rate: if the context of a word contains few errors,
then we are able compute expectations for that word confidently
(resulting in low surprisal). If the context contains lots of errors
then expecations are difficult to compute and they become unreliable
(resulting in high surprisal). We will now test these predictions
regarding error type and error rate using a character-based version of
surprisal.

\subsection{Methods}

We trained a character-based neural language model using LSTM cells
\cite{hochreiter_long_1997}.  Such models can assign probabilities to
any sequence of characters, and thus are capable of computing
surprisal even for words never seen in the training data, such as
erroneous words.  For training, we used the Daily Mail portion of the
DeepMind corpus. We used a vocabulary consisting of the 70 most
frequent characters, mapping others to an out-of-vocabulary token.

The hyperparameters of the language model were selected on an English
corpus based on Wikipedia text.\footnote{1024 units, 3 layers, batch
  size 128, embedding size 200, learning rate 3.6 with plain SGD,
  multiplied by 0.95 at the end of each epoch; BPTT length 80;
  DropConnect with rate 0.01 for hidden units; replacing entire
  character embeddings by zero with rate 0.001.}  We then used the
resulting model to compute surprisal on the texts used in the
eye-tracking experiment for each experimental condition.

The model estimates, for each element of a character sequence, the
probability of seeing this character given the preceding context.  We
compute the surprisal of a word as the sum of the surprisals of the
individual characters, as prescribed by the product rule of
probability. For a word consisting of characters $x_{t}\dots x_{t+T}$
following a context $x_1...x_{t-1}$, its surprisal is:
\begin{equation}
    -\log P(x_{t}\dots x_{t+T}|x_1...x_{t-1}) = \sum_{i=t}^{t+T} -\log P(x_i|x_{1}...x_{i-1})
\end{equation}
In this computation, we take whitespace characters to belong to the
preceding word.

To control for the impact of the random initialization of the neural
network at the beginning of training, we trained seven models with
identical settings but different random initializations.

The quality of character-based language models is conventionally
measured in Bits Per Character (BPC), which is the average surprisal,
to the base 2, of each character. On held-out data, our model achieves
a mean BPC value of 1.28 (SD 0.025), competitive with BPC values
achieved by state-of-the-art systems of similar datasets
(e.g.,~\citeNP{merity2018analysis}, report a BPC value of 1.23 on
Wikipedia text).

In the introduction we predicted that word-based surprisal is not able
to model the reading time pattern we found in our eye-tracking
experiment. In order to test this prediction, we compare our
character-level surprisal model to surprisal computed using a
conventional word-based neural language model.  Word-based models have
a fixed vocabulary, consisting of the most common words in the
training data; a typical vocabulary size is 10,000.  Words that were
not seen in the training data, and rare words, are represented by a
special out-of-vocabulary (OOV) token. From a cognitive perspective,
this corresponds to assuming that all unknown words (whether they
contain errors or not) are treated in the same way: they are
recognized as unknown, but not processed any further.  We used a
vocabulary size of 10,000. The hyperparameters of the word-based model
were selected on the same English Wikipedia corpus as the
character-based model.\footnote{1024 units, batch size 128, embedding
  size 200, learning rate 0.2 with plain SGD, multiplied by 0.95 at
  the end of each epoch; BPTT length 50; DropConnect with rate 0.2 for
  hidden units; Dropout 0.1 for input layer; replacing words by random
  samples from the vocabulary with rate 0.01 during training.}

\subsection{Results and Discussion}


In this section, we show that surprisal computed by a character-level
neural language model (\textsc{CharSurprisal}) is able to account for
the effects of errors on reading observed in our eye-tracking
experiments. We compute character-based surprisal for the texts used
in our experiments, and expect to obtain mean surprisal scores for
each experimental condition that resemble mean reading times. We will
also verify our prediction that word-based surprisal
(\textsc{WordSurprisal}) is not able to account for the effects
observed in our experimental data, due to the way it treats unknown
words.

\begin{figure}[tb]
    \centering
    \includegraphics[width=\columnwidth]{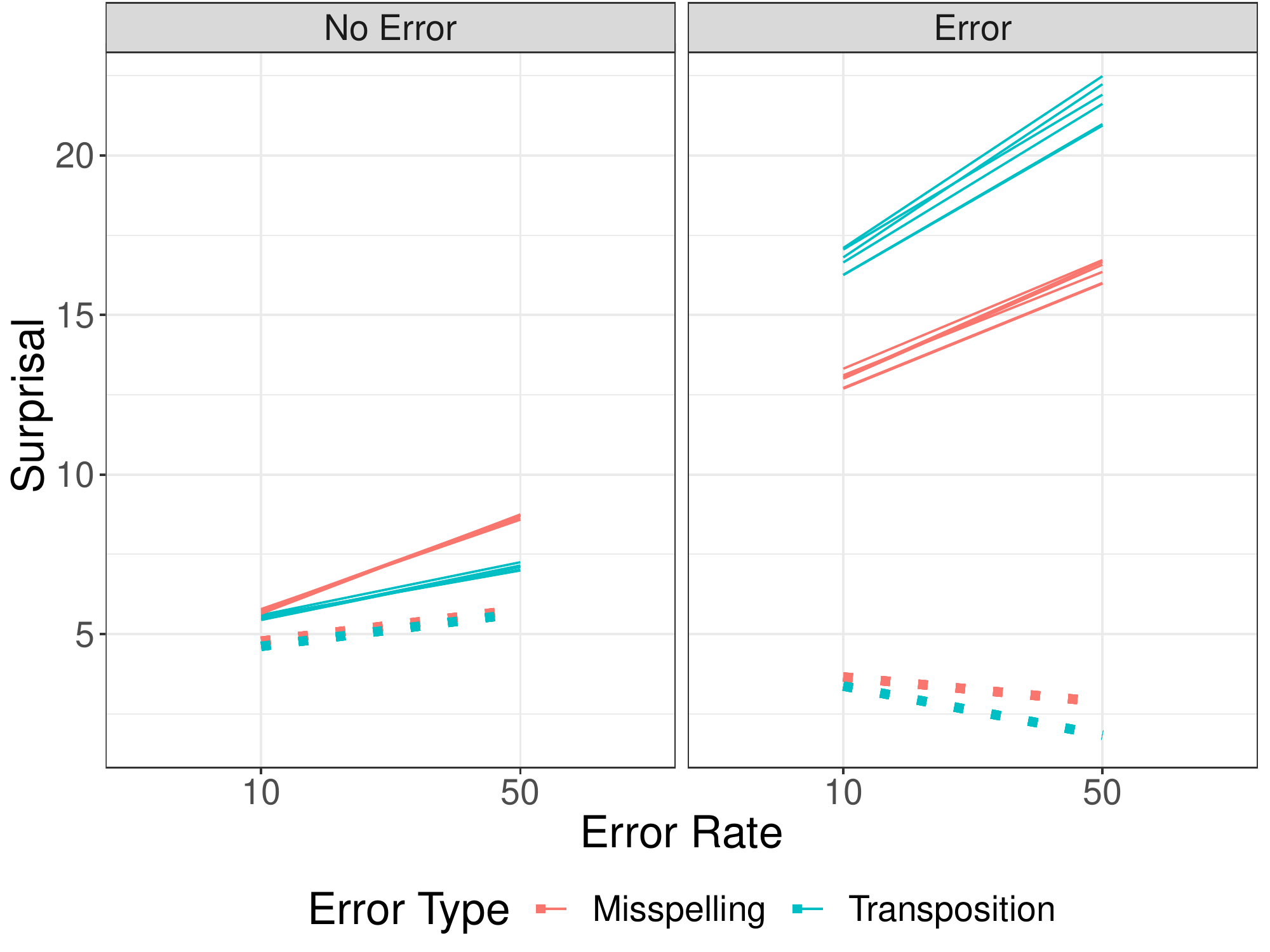}
    \caption{\textsc{CharSurprisal} (full lines) and
      \textsc{WordSurprisal} (dotted lines) as a function of error
      type and error rate, for correct (left) and erroneous (right)
      words. For \textsc{CharSurprisal}, we show the means of all
      seven random initializations of our neural surprisal model.}
    \label{fig:surp}
\end{figure}

Figure~\ref{fig:surp} shows the mean surprisal values across the
different error conditions. We note that the pattern of reading time
predicted by \textsc{CharSurprisal} (solid lines) matches the
first-pass times observed experimentally very well (see
Figure~\ref{fig:fp}), while \textsc{WordSurprisal} (dotted line) shows
a clearly divergent pattern, with error words showing \emph{lower}
surprisal than non-error words. This can be explained by the fact that
a word-based model does not process error words beyond recognizing
them as unknown; the presence of an unknown word itself is not a
high-surprisal event (even without errors, 17~\% of the words in our
texts are unknown to the model, given its 10,000-word vocabulary).

To confirm this observation statistically, we fitted linear
mixed-effects models with \textsc{CharSurprisal} and
\textsc{WordSurprisal} as dependent variables.  We enter the seven
random initializations of each model as a random factor, analogously
to the participants in the eye-tracking experiment.  We use the same
predictors that we used for the reading measures, except for
\textsc{LastFix}. This predictor is not available: suprisal models
compute a difficulty measure for each word (viz.,~its surprisal), but
they are not able to predict whether a word will be skipped or not.

The results of the mixed model with \textsc{CharSurprisal} as the
dependent variable (see Table~\ref{tab:mixed-models-surp}) replicated
the effects of \textsc{ErrorRate}, \textsc{Error}, and
\textsc{WordLength} found in first pass and fixation rate, as well as
the effect of \textsc{ErrorType} found only in fixation rate (see
Table~\ref{tab:mixed-models}). The same mixed model with
\textsc{WordSurprisal} as the dependent variable (see again
Table~\ref{tab:mixed-models-surp}), however, does not yield the
correct pattern of results: Crucially, the coefficients of
\textsc{Error} and \textsc{ErrorType} have the opposite sign compared
to both \textsc{CharSurprisal} and the experimental data (though both
effects are small, see dotted lines in Figure~\ref{fig:surp}).

\begin{table}[tb]
\centering
\begin{tabular}{@{~}l@{~~~}d@{}d@{}l@{~~~}d@{}d@{}l@{~}}
\toprule
& \mc{3}{c}{\textsc{CharSurpr}} & \mc{3}{c}{\textsc{WordSurpr}} \\
\midrule
(Intercept)                & 10.47   & (0.09) & $^{***}$  & 5.06 & (0.07) &  $^{***}$\\
\textsc{ErrType}                 & 1.27 & (0.02) & $^{***}$  & -0.40 & (0.02)& $^{***}$\\
\textsc{ErrRate}                 &  1.57 & (0.02) & $^{***}$ &0.01& (0.00)& $^{***}$\\
\textsc{Error}                      & 13.88 & (0.03) & $^{***}$ &-2.96&(0.02)& $^{***}$\\
\textsc{WLength}                &  3.02 & (0.05) & $^{***}$ &0.25&0.01& $^{***}$\\
\bottomrule
\mc{7}{c}{$Pr(\beta < 0)$:  $^{***}<0.001$, $^{**}<0.01$, $^*<0.05$}
\end{tabular}
\caption{Models of character-level and word-level
  surprisal with random effects for model runs and items. Each cell
  gives the coefficient, its standard deviation and the estimated
  posterior probability that the coefficient has the opposite
  sign.}\label{tab:mixed-models-surp}
\end{table}

\begin{table}[tb]
\centering
\begin{tabular}{@{~}l@{}d@{}d@{}l@{~}d@{}d@{}l@{~}}
\toprule
& \mc{3}{c}{First Pass}& \mc{3}{c}{Fixation Rate} \\
\midrule
(Intercept)                & 248.73 & (5.55) & $^{***}$& -0.15   & (0.09) & $^{}$  \\
\textsc{WLength}                &  22.22 & (0.79) & $^{***}$& 0.75  & (0.01)  & $^{***}$  \\
\textsc{LastFix}                    & 2.65   & (1.34) &         & 0.22 & (0.02) & $^{***}$ \\ 
\textsc{WLength} $\times$ \textsc{LastFix} & \mc{3}{c}{---}        & 0.60  & (0.19) & $^{***}$ \\
\bottomrule
\textsc{ResidCharSurp-} & 9.89 & (0.78)   &  $^{***}$   &  0.09 & (0.01) & $^{***}$ \\ 
\textsc{Oracle}\\
\midrule
\textsc{ResidCharSurp}   & 13.82    & (0.66) &  $^{***}$   &  0.14 & (0.01) & $^{***}$ \\
\bottomrule
$\Delta$AIC                    & \mc{3}{c}{$-273.88$}             & \mc{3}{c}{$-205.83$} \\
$\Delta$BIC                    & \mc{3}{c}{$-273.88$}             & \mc{3}{c}{$-205.83$} \\
\bottomrule
\mc{7}{c}{$Pr(\beta < 0)$:  $^{***}<0.001$, $^{**}<0.01$, $^*<0.05$}
\end{tabular}
\caption{Models for reading measures with surprisal predictors. We
  compare model fit between a model with character-based
  surprisal (\textsc{ResidCharSurp}) and character-based
  oracle surprisal (\textsc{ResidCharSurpOracle}), both residualized
  against word length.}\label{tab:mixed-models-plus-surp}
\end{table}

We have shown that character-based surprisal computed on the texts
used in our experiment is qualitatively similar to the experimental
results. As a next step we will test its quantitative predictions,
i.e., we will correlate surprisal scores with reading times. For this,
we performed mixed-effects analyses in which first-pass time and
fixation rate are predicted by \textsc{WLength}, \textsc{LastFix}, and
character-based surprisal residualized against word length
(\textsc{ResidCharSurp}).\footnote{The correlation between word length
  and raw surprisal is 0.26.} Note that we did not enter the error
factors (\textsc{ErrorType}, \textsc{ErrorRate}, \textsc{Error}) into
this analysis, as we predict that surprisal will simulate the effect
of errors in reading.

It is known that surprisal predicts reading times in ordinary text not
containing errors~\cite{demberg_data_2008,Frank:09}; thus, it is
important to disentangle the specific contribution of modeling errors
correctly from the general contribution of surprisal in our model. We
do this by constructing a baseline version of character-based
surprisal that is computed using an oracle
(\textsc{ResidCharSurpOracle}). For this, we replace erroneous words
with their correct counterparts before computing surprisal, and again
residualize against word length.\footnote{The correlation between word
  length and unresidualized oracle surprisal is 0.47.}  If
\textsc{ResidCharSurp} correctly accounts for the effects of errors on
reading, then we expect that \textsc{ResidCharSurp} -- which has
access to the erroneous word forms -- will improve the fit with our
reading data compared to \textsc{ResidCharSurpOracle}.

For \textsc{ResidCharSurpOracle}, we use the same seven models as for
\textsc{ResidCharSurp}, only exchanging the character sequences on
which surprisal is computed.  This ensures that any difference in
model fit between the two predictors can be attributed entirely to the
way \textsc{ResidCharSurp} is affected by the presence of errors in
the texts.

The resulting models are shown in
Table~\ref{tab:mixed-models-plus-surp}.  For \textsc{WLength} and
\textsc{LastFix}, we see the same pattern of results as in the
experimental data (see Table~\ref{tab:mixed-models}). Furthermore,
regular surprisal (\textsc{ResidCharSurp}) and oracle surprisal
(\textsc{ResidCharSurpOracle}) significantly predict both first pass
time and fixation rate. This is in line with the standard finding that
surprisal predicts reading time \cite{demberg_data_2008,Frank:09}, but
has so far not been demonstrated for texts containing errors. We
compare model fit using AIC and BIC. Both measures indicate that
\textsc{ResidCharSurp} fits the experimental data better than
\textsc{ResidCharSurpOracle}.  Thus, character-level surprisal
provides an account of our data going beyond the known contribution of
ordinary surprisal to reading times, and correctly predicts reading in
the presence of errors.

\section{Conclusion}

We investigated reading with errors in texts that contain either
letter transpositions or real misspellings. We found that
transpositions cause more reading difficulty than misspellings and
explained this using a character-based surprisal model, which assigns
higher surprisal to rare letter sequences as they occur in
transpositions. We also found that in texts with a high error rate,
all words are more difficult to read, even the ones without
errors. Again, character-based surprisal explains this: computing word
expectations is harder when the context of a word is degraded by
errors, resulting in increased surprisal.

In future work, we plan to integrate character-based surprisal with
existing neural models of human reading \cite{Hahn:Keller:18}. Models
at the character level are necessary not only to account for errors,
but also to model landing position effects, parafoveal preview, and
word length effects, all of which word-based models are unable to
capture.

\section{Acknowledgements}
Y.B.\ was supported by the Harvard Mind, Brain, and Behavior
Initiative.  F.K.\ was supported by the Leverhulme Trust through
International Academic Fellowship IAF-2017-019.

\bibliographystyle{apacite}

\setlength{\bibleftmargin}{.125in}
\setlength{\bibindent}{-\bibleftmargin}

\bibliography{references}

\end{document}